\pdfoutput=1

\documentclass[11pt]{article}

\usepackage[]{EMNLP2022}

\usepackage{times}
\usepackage{latexsym}
\usepackage{amsfonts}
\usepackage{graphicx}
\usepackage{svg}
\usepackage{multirow}
\usepackage{bm}
\usepackage[T1]{fontenc}

\usepackage[utf8]{inputenc}

\usepackage{microtype}

\title{MLKD-BERT: Multi-level Knowledge Distillation for Pre-trained Language Models}

\author{Ying Zhang$^{1}$, Ziheng Yang$^{1}$, Shufan Ji$^{1}$\\
\textsuperscript{\rm1}Beihang University\\
{\texttt{\{yingzhang1998, yzh2206140, jishufan\}@buaa.edu.cn}}}

\begin{document}
\maketitle
\begin{abstract}
Knowledge distillation is an effective technique for pre-trained language model compression.
Although existing knowledge distillation methods perform well for the most typical model BERT, they could be further improved in two aspects: the relation-level knowledge could be further explored to improve model performance; and the setting of student attention head number could be more flexible to decrease inference time. 
Therefore, we are motivated to propose a novel knowledge distillation method MLKD-BERT to distill multi-level knowledge in teacher-student framework.
Extensive experiments on GLUE benchmark and extractive question answering tasks demonstrate that our method outperforms state-of-the-art knowledge distillation methods on BERT. In addition, MLKD-BERT can flexibly set student attention head number, allowing for substantial inference time decrease with little performance drop. 
\end{abstract}

\section{Introduction}
In recent years, large-scale pre-trained language models (PLMs) have been widely applied in natural language processing, such as
BERT \citep{DBLP:conf/naacl/DevlinCLT19}, 
XLNet \citep{DBLP:conf/nips/YangDYCSL19}, 
and RoBERTa \citep{DBLP:journals/corr/abs-1907-11692}.
The PLM usually has large number of parameters and long inference time, making it inapplicable to resource-limited devices and real-time scenarios.
Therefore, it is crucial to reduce PLM's storage and computation overhead while retaining its performance. 
Knowledge distillation \citep{DBLP:journals/corr/HintonVD15} is an effective technique for PLM compression.
In knowledge distillation, a smaller compact student model is trained, under the guidance of a larger complicated teacher model, to keep similar model performance.

As for the most typical PLM BERT, there exist several knowledge distillation methods for model compression, including
DistilBERT \citep{DBLP:journals/corr/abs-1910-01108},
BERT-PKD \citep{DBLP:conf/emnlp/SunCGL19},
TinyBERT \citep{DBLP:conf/emnlp/JiaoYSJCL0L20},
BERT-EMD \citep{DBLP:conf/emnlp/LiLZXYJ20},
M\small{INI}\normalsize LM \citep{DBLP:conf/nips/WangW0B0020},
and M\small{INI}\normalsize LMv2 \citep{DBLP:conf/acl/WangBHDW21}.  
Despite the effectiveness of previous methods, there still exist two problems not addressed well:

\begin{itemize}
\item
Existing methods mainly distill feature-level knowledge, but seldom consider relation-level knowledge (relation among tokens and relation among samples). However, the relation-level knowledge may be valuable to improve the performance of student model.
\item
Most previous works use self-attention distribution to distill teacher’s self-attention modules, thus the student model is restricted to take the same attention head number as its teacher. Such restriction prevents the reduction of attention head number in student model, resulting in increased inference time.
\end{itemize}
Therefore, a more flexible knowledge distillation method with improved performance is preferred to transfer knowledge from teacher to student model.

In this paper, we propose a novel multi-level knowledge distillation method MLKD-BERT for BERT compression. 
MLKD-BERT conducts a two-stage distillation for feature-level as well as relation-level knowledge, with 6 distillation loss functions designed for embedding layer, Transformer layers, and prediction layer.
Compared with previous works, we have made two main contributions:
\begin{itemize}
\item
In addition to feature-level knowledge, our student model learns valuable relation-level knowledge (relation among tokens and relation among samples) from its teacher model, which further improves the performance.

\item Our student model learns self-attention relation instead of self-attention distribution, making it flexible in attention head number setting. As such, our student model could reduce attention head number to further decrease inference time. 
\end{itemize}

Extensive experiments on GLUE \citep{DBLP:conf/iclr/WangSMHLB19} benchmark and extractive question answering tasks show that our MLKD-BERT outperforms state-of-the-art BERT distillation methods on various prediction tasks. 
In addition, MLKD-BERT can set smaller student attention head number, allowing for substantial inference time decrease with little performance drop. Moreover, MLKD-BERT is effective in PLM compression, in that MLKD-BERT keeps competitive performance ($99.5\%$ on average for GLUE tasks) as its teacher with $50\%$ compression in parameters and inference time. 

The rest of the paper is organized as follows. Related works are reviewed in Section \ref{section:sec2}. We introduce our method MLKD-BERT in Section \ref{section:sec4}, and conduct extensive experiments in Section \ref{section:sec5}. Finally, in Section \ref{section:sec6}, conclusions are drawn.

\section{Related Works}\label{section:sec2}

\subsection{Pre-trained Language Models}
Nowadays, large-scale pre-trained language models have significantly improved the performance of many natural language processing tasks. Pre-trained language models are usually trained on large amounts of text data, and then fine-tuned for specific task. Early research efforts mainly focus on word embedding, such as word2vec \citep{DBLP:conf/nips/MikolovSCCD13} and GloVe \citep{DBLP:conf/emnlp/PenningtonSM14}.
Subsequently, researchers have shifted to contextual word embedding, including BERT \citep{DBLP:conf/naacl/DevlinCLT19}, GPT \cite{radford2018improving}, ENRIE \citep{DBLP:conf/acl/ZhangHLJSL19}, XLNet \citep{DBLP:conf/nips/YangDYCSL19} and RoBERTa \citep{DBLP:journals/corr/abs-1907-11692}.
However, those PLMs contain millions of parameters and take long inference time, making them inapplicable to resource-limited devices and real-time scenarios.
Fortunately, there exist many compression techniques for PLMs, which reduce model size and accelerate model inference while keeping model performance.

\subsection{Knowledge Distillation}
Besides quantization \citep{DBLP:conf/aaai/ShenDYMYGMK20} and network pruning \citep{DBLP:conf/emnlp/WangWL20}, knowledge distillation \citep{DBLP:journals/corr/abs-1903-12136} has been proven to be an effective technique for PLM compression.
As for the most typical PLM BERT, there exist several knowledge distillation methods for model compression.
Distilled BiLSTM \citep{DBLP:journals/corr/abs-1903-12136} tries to distill knowledge from BERT into a simple LSTM.
DistilBERT \citep{DBLP:journals/corr/abs-1910-01108} uses soft target probabilities and embedding outputs to train student model. 
BERT-PKD \citep{DBLP:conf/emnlp/SunCGL19} learns from multiple intermediate layers of teacher model for incremental knowledge extraction. 
TinyBERT \citep{DBLP:conf/emnlp/JiaoYSJCL0L20}, MobileBERT \citep{DBLP:conf/acl/SunYSLYZ20}, and SID \citep{DBLP:conf/aaai/AguilarLZYFG20} further improve BERT-PKD by distilling more internal representations, such as embedding layer outputs and self-attention distribution.
BERT-EMD \citep{DBLP:conf/emnlp/LiLZXYJ20} allows each intermediate student layer to learn from any intermediate teacher layer based on Earth Mover’s Distance.
M\small{INI}\normalsize LM \citep{DBLP:conf/nips/WangW0B0020} uses self-attention distribution and value relation to conduct deep self-attention distillation. 
M\small{INI}\normalsize LMv2 \citep{DBLP:conf/acl/WangBHDW21} generalizes deep self-attention distillation in M\small{INI}\normalsize LM \citep{DBLP:conf/nips/WangW0B0020}, employing self-attention relation.

Although existing knowledge distillation methods perform well in BERT compression, they are limited in two aspects: the relation-level knowledge is not well explored by the student model to enhance performance; and the attention head number of student model is restricted to the same as its teacher, increasing the inference time. Hence, we are motivated to propose a more flexible knowledge distillation method for BERT with improved performance.

\section{Method}\label{section:sec4}
\begin{figure}[ht]
\centering
\includegraphics[scale=0.5]{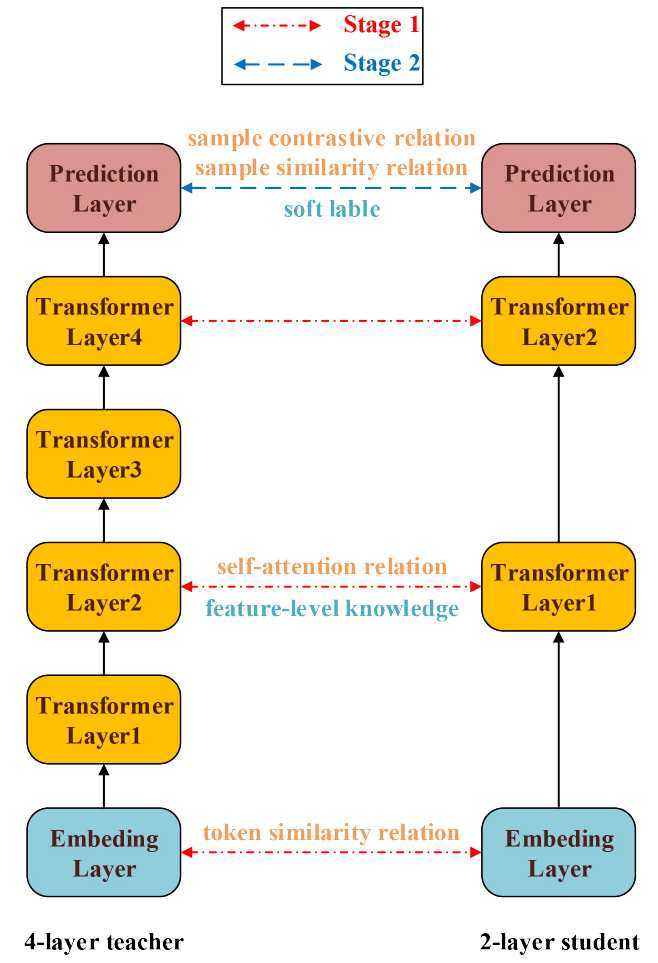}
\caption{Framework of MLKD-BERT.}
\label{fig:fig1}
\end{figure}

MLKD-BERT employs a two-stage distillation procedure for downstream task prediction in teacher-student framework (illustrated in Figure \ref{fig:fig1}).
Stage 1 distills embedding-layer and Transformer-layers, emphasizing 
feature representation and transformation, while Stage 2 distills prediction-layer, emphasizing sample prediction. 
In knowledge distillation, each student layer is mapped to corresponding teacher layer. 
As the number of Transformer-layers in student model is smaller than that in teacher model, we take uniform mapping strategy \citep{DBLP:conf/emnlp/JiaoYSJCL0L20} for Transformer-layer (TL) mapping.
For example, in Figure~\ref{fig:fig1}, when 2 student TLs are mapped to 4 teacher TLs, TL1 and TL2 in student model are mapped to TL2 and TL4 in teacher model, respectively.

In MLKD-BERT, we bring in some new relation-level knowledge to improve layer distillation. 
At embedding-layer, token similarity relation is employed to enhance feature representation. 
At Transformer-layers, besides feature-level knowledge, self-attention relation is proposed for flexible student attention head number setting. 
At prediction-layer, in addition to soft labels, sample similarity relation and sample contrastive relation are introduced to enhance prediction. 
The distillation procedures will be detailed in the following subsections.

\subsection{Embedding-layer Distillation}
Since embedding-layer performs feature representation of tokens for each data sample, token similarity relation could be valuable knowledge to enhance embedding-layer distillation.
In embedding-layer distillation, token similarity relation is transferred from teacher to student by minimizing the KL-divergence of token embedding similarities between teacher and student, according to the embedding distillation loss function $\mathcal{L}_{\mbox{\scriptsize{EMB}}}$ defined in Eqn.(\ref{eqa:eqa8}):

\begin{equation}
    \mathcal{L}_{\mbox{\scriptsize{EMB}}}=\frac{1}{|x|}\sum_{i=1}^{|x|}D_{\mbox{\scriptsize{KL}}}(\mathbf{R}_{i}^{T}||\mathbf{R}_{i}^{S})\label{eqa:eqa8}
\end{equation}
\begin{equation}
    \mathbf{R}^{T}=\mbox{softmax}(\frac{\mathbf{E}^T\mathbf{E}^{T\mathsf{T}}}{\sqrt{d_h^T}})
\end{equation}
\begin{equation}
    \mathbf{R}^{S}=\mbox{softmax}(\frac{\mathbf{E}^S\mathbf{E}^{S\mathsf{T}}}{\sqrt{d_h^S}})
\end{equation}
where $|x|$ is the length of input sequence;
matrices $\mathbf{E}^T\in\mathbb{R}^{|x|\times{d_h^T}}$ and $\mathbf{E}^S\in\mathbb{R}^{|x|\times{d_h^S}}$ are token embeddings of teacher and student;
$d_h^T$ and $d_h^S$ are hidden dimension of teacher and student;
matrices $\mathbf{R}^T\in\mathbb{R}^{|x|\times{|x|}}$ and $\mathbf{R}^S\in\mathbb{R}^{|x|\times{|x|}}$ are token embedding similarity matrices of teacher and student, respectively.

\subsection{Transformer-layer Distillation}
\begin{figure}[ht]
\centering
\includegraphics[scale=0.4]{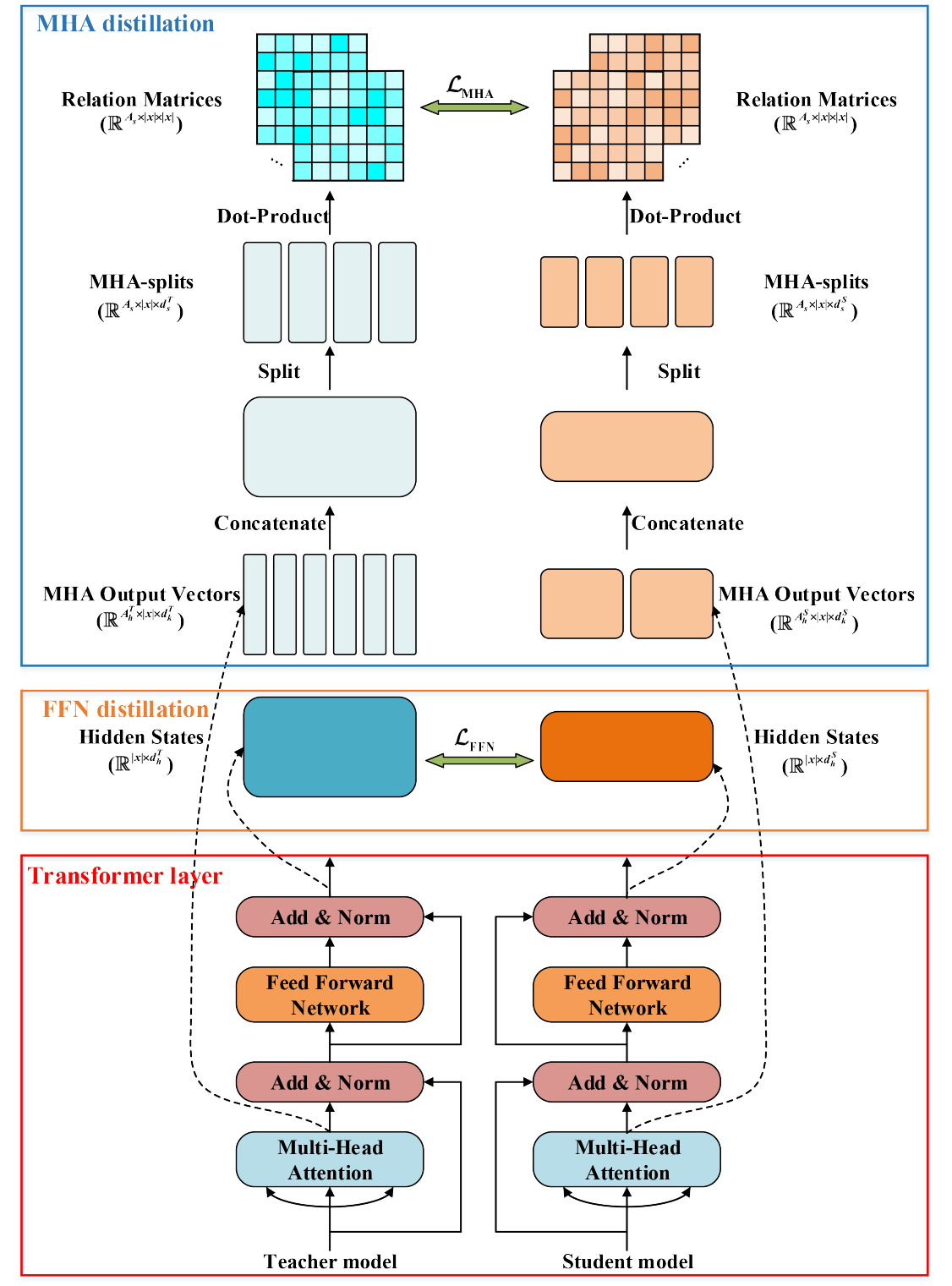}
\caption{Transformer-layer Distillation: MHA Distillation and FFN Distillation.}
\label{fig:fig2}
\end{figure}

Standard Transformer-layer contains two main sub-layers: Multi-Head Attention (MHA) and Feed Forward Network (FFN).
Transformer-layer distillation (illustrated in Figure \ref{fig:fig2}) is conducted on MHA and FFN to transfer self-attention relation and feature-level knowledge, respectively.

As for MHA sub-layer, the similarities among its output vectors are defined as self-attention relation.
At Transformer layer $l$, let $A_h$ represents the number of attention heads, then output $\mathbf{O}_{l,a} (a\in[1,A_h]$) of the $a$-th attention head, is computed via:
\begin{equation}
    \mathbf{O}_{l,a}=\mbox{softmax}(\frac{\mathbf{Q}_{l,a}\mathbf{K}_{l,a}^{\mathsf{T}}}{\sqrt{d_k}})\mathbf{V}_{l,a}\label{eqa:eqa5}
\end{equation}
\begin{equation}
    \mathbf{Q}_{l,a}=\mathbf{H}^{l-1}\mathbf{W}_{l,a}^{Q}
\end{equation}
\begin{equation}
    \mathbf{K}_{l,a}=\mathbf{H}^{l-1}\mathbf{W}_{l,a}^{K}
\end{equation}
\begin{equation}
    \mathbf{V}_{l,a}=\mathbf{H}^{l-1}\mathbf{W}_{l,a}^{V}
\end{equation}
where $\mathbf{H}^{l-1}\in\mathbb{R}^{|x|\times{d_h}}$ is the input vectors of Transformer layer $l-1$, with $|x|$ representing the length of input sequence and $d_h$ representing the hidden dimension;
$\mathbf{Q}_{l,a}$, $\mathbf{K}_{l,a}$, $\mathbf{V}_{l,a}$ are linearly projections of $\mathbf{H}^{l-1}$;
$\mathbf{W}_{l,a}^Q, \mathbf{W}_{l,a}^K, \mathbf{W}_{l,a}^V\in\mathbb{R}^{d_h\times{d_k}}$ are parameter matrices;
and $d_k$ is attention head size.

In MHA distillation, MHA outputs (i.e., $\textbf{O}_{l,a}$ defined in Eqn.(\ref{eqa:eqa5})) are concatenated together and then split into a certain number of vector groups (named MHA-splits), in both teacher and student models. 
We suggest setting the number of MHA-splits as the number of student attention heads.
After that, self-attention relation is transferred from teacher to student by minimizing the KL-divergence of MHA output vector similarities between teacher and student. Given that the $n$-th student layer is mapped to the $m$-th teacher layer, the loss function for MHA distillation $\mathcal{L}_{\mbox{\scriptsize{MHA}}}$ is defined in Eqn.(\ref{eqa:eqa11}):

\begin{equation}
    \mathcal{L}_{\mbox{\scriptsize{MHA}}}=\frac{1}{A_s|x|}\sum_{n=1}^{N}\sum_{a=1}^{A_s}\sum_{i=1}^{|x|}D_{\mbox{\scriptsize{KL}}}(\mathbf{R}_{m,a,i}^{T}||\mathbf{R}_{n,a,i}^{S})\label{eqa:eqa11}
\end{equation}
\begin{equation}
    \mathbf{R}_{m,a}^{T}=\mbox{softmax}(\frac{\mathbf{O}_{m,a}^T\mathbf{O}_{m,a}^{T\mathsf{T}}}{\sqrt{d_s^T}})
\end{equation}
\begin{equation}
    \mathbf{R}_{n,a}^{S}=\mbox{softmax}(\frac{\mathbf{O}_{n,a}^S\mathbf{O}_{n,a}^{S\mathsf{T}}}{\sqrt{d_s^S}})
\end{equation}
where $|x|$ is the length of input sequence;
$A_s$ is the number of MHA-splits;
$N$ is the number of student Transformer-layers;
matrices $\mathbf{O}_{m,a}^T\in\mathbb{R}^{|x|\times{d_s^T}}$ and $\mathbf{O}_{n,a}^S\in\mathbb{R}^{|x|\times{d_s^S}}$ are MHA outputs in MHA-split $a$ at teacher's Layer $m$ and student's Layer $n$;
$d_s^T$ and $d_s^S$ are split-head size of teacher and student MHA-split;
matrices $\mathbf{R}_{m,a}^{T}\in\mathbb{R}^{|x|\times{|x|}}$ and 
$\mathbf{R}_{n,a}^{S}\in\mathbb{R}^{|x|\times{|x|}}$ are MHA output vector similarities in MHA-split $a$ at teacher's Layer $m$ and student's Layer $n$, respectively.

Note that the teacher and student model should have the same number of MHA-splits.
In MLKD-BERT, self-attention relation is transferred by MHA-split rather than MHA attention head.
As such, student model does not have to set the same number of attention heads as its teacher.
In this way, the number of attention heads in the student model can be set smaller, which could decrease inference time. 

As for FFN sub-layer, the feature-level knowledge is distilled by minimizing the mean squared error of the output hidden states between teacher and student, according to the FFN distillation loss function $\mathcal{L}_{\mbox{\scriptsize{FFN}}}$ in Eqn.(\ref{eqa:eqa14}):

\begin{equation}
    \mathcal{L}_{\mbox{\scriptsize{FFN}}}=\sum_{n=1}^{N}\mbox{MSE}(\mathbf{H}_n^S\mathbf{W}_h,\mathbf{H}_{m}^T)\label{eqa:eqa14}
\end{equation}
Given that the $n$-th student layer is mapped to the $m$-th teacher layer, matrices $\mathbf{H}_n^S\in\mathbb{R}^{|x|\times{d_h^S}}$ and $\mathbf{H}_{m}^T\in\mathbb{R}^{|x|\times{d_h^T}}$ are the hidden states of student's Layer $n$ and teacher's Layer $m$; 
matrix $\mathbf{W}_h\in\mathbb{R}^{d_h^S\times{d_h^T}}$ is a learnable linear transformation, which transforms the student's hidden states into the same space as the teacher’s hidden states.

In summary, as the first stage distillation covers both embedding-layer and Transformer-layers, the above distillation loss functions are summed up to the first stage distillation loss function in Eqn.(\ref{eqa:eqa15}):

\begin{equation}
    \mathcal{L}_{\mbox{\scriptsize{Stage\mbox{\,}1}}}=\mathcal{L}_{\mbox{\scriptsize{EMB}}}+\mathcal{L}_{\mbox{\scriptsize{MHA}}}+\mathcal{L}_{\mbox{\scriptsize{FFN}}}\label{eqa:eqa15}
\end{equation}

\subsection{Prediction-layer Distillation}

As samples with same class label tend to be similar than samples with different class labels, relation among samples would be valuable knowledge for prediction-layer distillation. Here, we will bring in sample similarity relation and sample contrastive relation to enhance prediction-layer distillation.

Sample similarity relation is defined as the sample similarities within a data batch, without considering sample labels. The sample similarity relation is distilled from teacher to student by minimizing the KL-divergence of sample similarities between teacher and student, according to the sample-similarity distillation loss function $\mathcal{L}_{\mbox{\scriptsize{SS}}}$ defined in Eqn.(\ref{eqa:eqa16}):

\begin{equation}
    \mathcal{L}_{\mbox{\scriptsize{SS}}}=\frac{1}{b}\sum_{i=1}^{b}D_{\mbox{\scriptsize{KL}}}(\mathbf{R}_{i}^{T}||\mathbf{R}_{i}^{S})\label{eqa:eqa16}
\end{equation}
\begin{equation}
    \mathbf{R}^{T}=\mbox{softmax}(\frac{\mathbf{G}^T\mathbf{G}^{T\mathsf{T}}}{\sqrt{d_h^T}})\label{eqa:eqa17}
\end{equation}
\begin{equation}
    \mathbf{R}^{S}=\mbox{softmax}(\frac{\mathbf{G}^S\mathbf{G}^{S\mathsf{T}}}{\sqrt{d_h^S}})\label{eqa:eqa18}
\end{equation}
where $b$ is batch size;
$d_h^T$ and $d_h^S$ are hidden dimension of teacher and student;
matrices $\mathbf{G}^T\in\mathbb{R}^{b\times{d_h^T}}$ and $\mathbf{G}^S\in\mathbb{R}^{b\times{d_h^S}}$ are sample representations in a batch, i.e., [CLS] outputs from the last Transformer-layer of teacher and student;
$\mathbf{R}^{T}\in\mathbb{R}^{b\times{b}}$ and $\mathbf{R}^{S}\in\mathbb{R}^{b\times{b}}$ are sample similarity matrices of teacher and student, respectively.

Sample contrastive relation is employed to map samples with same and different class labels (according to ground truth of training samples) into close and distant representation space, respectively.  
Sample contrastive relation is distilled by minimizing the sample-contrastive distillation loss function $\mathcal{L}_{\mbox{\scriptsize{SC}}}$
in Eqn.(\ref{eqa:eqa19}) \citep{DBLP:conf/nips/KhoslaTWSTIMLK20}:

\begin{equation}
    \mathcal{L}_{\mbox{\scriptsize{SC}}}=\frac{1}{2b}\sum_{i=1}^{2b}\sum_{i\in{I}}\frac{1}{|P(i)|}\sum_{p\in{P(i)}}\mathcal{L}_{\mbox{\scriptsize{InfoNCE}}}(i,p)\label{eqa:eqa19}
\end{equation}
\begin{equation}
    \mathcal{L}_{\mbox{\scriptsize{InfoNCE}}}(i,p)=-\mbox{log}\frac{\mbox{exp}(\mathbf{h}_i\cdot\mathbf{h}_p)/\rho}{\sum_{a\in{A(i)}}\mbox{exp}(\mathbf{h}_i\cdot\mathbf{h}_a)/\rho}
\end{equation}
where $b$ is batch size; $I\equiv{\{1,...,2b\}}$, $A(i)\equiv{I\backslash\{i\}}$, $P(i)\equiv{\{p|p\in{A(i)}, y_p=y_i\}}$; $y_i$ is class label of $i$-th sample; $\rho$ is scalar temperature parameter; $\textbf{h}_i$ is the $i$-th row of $\mathbf{H}\in\mathbb{R}^{2b\times{d_h^T}}$; $\mathbf{H}=\mbox{Concat}(\mathbf{G}^S\mathbf{W}_g, \mathbf{G}^T)=[\textbf{h}_1;...;\textbf{h}_{2b}], i.e., \mathbf{G}^S\mathbf{W}_g=[\mathbf{h}_1;...;\mathbf{h}_b], \mathbf{G}^T=[\mathbf{h}_{b+1};...;\mathbf{h}_{2b}]$; $\mathbf{W}_g\in\mathbb{R}^{d_h^S\times{d_h^T}}$ is linear transformation matrix; $\mathbf{G}^T$ and $\mathbf{G}^S$ are defined in Eqn.(\ref{eqa:eqa17}) and Eqn.(\ref{eqa:eqa18}).  

Similar to previous distillation works, we also adopt soft label distillation by minimizing the soft-label distillation loss function $\mathcal{L}_{\mbox{\scriptsize{KD}}}$ in Eqn.(\ref{eqa:eqa20}):

\begin{equation}
    \mathcal{L}_{\mbox{\scriptsize{KD}}}=D_{\mbox{\scriptsize{KL}}}(\mbox{softmax}(\mathbf{z}^T/\tau)||\mbox{softmax}(\mathbf{z}^S/\tau))\label{eqa:eqa20}
\end{equation}
where $\tau$ is scalar temperature parameter;
$\mathbf{z}^T$ and $\mathbf{z}^S$ are logits predicted by teacher and student, respectively. 

\begin{table*}[t]
\centering
\begin{small}
\resizebox{\textwidth}{!}{
\begin{tabular}{l|cc|cccccccc|c}
\hline
\multirow{2}*{\textbf{Model}} & \multirow{2}*{\textbf{\#Params}} & \multirow{2}*{\textbf{Speedup}} & \textbf{MNLI-m/-mm} & \textbf{QQP} & \textbf{QNLI} & \textbf{SST-2} & \textbf{CoLA} & \textbf{STS-B} & \textbf{MRPC} &\textbf{RTE} &\multirow{2}*{\textbf{Avg}}\\
&&&(393k)&(364k)&(105k)&(67k)&(8.5k)&(5.7k)&(3.7k)&(2.5k)\\
\hline
BERT-base & 109M & $1.0\times$ & 84.2/83.6 & 71.6 & 90.8 & 94.3 & 52.6 & 83.9 & 87.3 & 67.3 & 79.5\\
\hline
$\mbox{BERT}_{\mbox{\tiny{TINY}}}$ & 14.5M &  $9.4\times$ & 75.4/74.9 & 66.5 & 84.8 & 87.6 & 19.5 & 77.1 & 83.2 & 62.6 & 70.2\\
$\mbox{BERT}_4\mbox{-PKD}$ & 52.2M & $3.0\times$ & 79.9/79.3 & 70.2 & 85.1 & 89.4 & 24.8 & 79.8 & 82.6 & 62.3 & 72.6\\
$\mbox{DistilBERT}_4$ &  52.2M & $3.0\times$ & 78.9/78.0 & 68.5 & 85.2 & 91.4 & 32.8 & 76.1 & 82.4 & 54.1 & 71.9\\
$\mbox{BERT-EMD}_4$ & 14.5M & $9.4\times$ & \textbf{82.1}/80.6& 69.3 & 87.2 & 91.0 & 25.6 & \textbf{82.3} & \textbf{87.6} & \textbf{66.2} & 74.7\\
$\mbox{TinyBERT}_4$ & 14.5M & $9.4\times$ & 81.4/80.4 & 69.9 & 85.9 & \textbf{91.9} & 35.2 & 81.5 & 85.4 & 62.1 & 74.8\\
$\bm{\mbox{MLKD-BERT}_4}$ & 14.5M & $9.4\times$ & 82.0/\textbf{80.7} & \textbf{70.6} & \textbf{87.5} & \textbf{91.9} & \textbf{35.5} & 81.9 & 86.3 & 63.5 & \textbf{75.6}\\
\hline
$\mbox{BERT}_6\mbox{-PKD}$ & 67.0M & $2.0\times$ & 81.5/81.0 & 70.7 & 89.0 & 92.0 & 43.5 & 81.6 & 85.0 & 65.5 & 76.6\\
$\mbox{DistilBERT}_6$ & 67.0M & $2.0\times$ & 82.6/81.3 & 70.1 & 88.9 & 92.5 & \textbf{49.0} & 81.3 & 86.9 & 58.4 & 76.8\\
$\mbox{TinyBERT}_6$ & 67.0M & $2.0\times$ & 83.9/83.4 & 72.0 & 89.9 & \textbf{93.7} & 46.7 & 83.3 & 85.7 & 66.6 & 78.4\\
M\scriptsize{INI}\small LMv2 & 67.0M & $2.0\times$ & 83.8/83.3 & 70.9 & 90.2 & 92.9 & 46.6 & \textbf{84.3} & \textbf{89.1} & \textbf{69.2} & 78.9 \\   
$\bm{\mbox{MLKD-BERT}_6}$ &  67.0M & $2.0\times$ & \textbf{84.4}/\textbf{83.5} & \textbf{72.2} & \textbf{90.8} & 93.3 & 48.0 & \textbf{84.3} & 87.3 & 67.8 & \textbf{79.1}\\
\hline
\end{tabular}
}
\end{small}
\caption{Comparative Studies on GLUE Benchmark. 
The number under each task represents the number of its training samples. 
Avg represents the average score over all tasks. 
The subscript within each model name represents the number of Transformer layers. 
The best result on each task is in-bold.
}
\label{tab:tab1}
\end{table*}

\begin{table}[t]
\centering
\begin{small}
\resizebox{\hsize}{!}{
\begin{tabular}{l|cc|c}
\hline
\multirow{2}*{\textbf{Model}}&\multirow{2}*{\textbf{SQuAD 1.1}}&\multirow{2}*{\textbf{SQuAD 2.0}}&\multirow{2}*{\textbf{Avg}}\\
&&&\\
\hline
BERT-base & 88.5 & 77.0 & 82.8\\
\hline
$\mbox{BERT}_4\mbox{-PKD}$ & 79.5 & 64.6 & 72.1 \\
$\mbox{DistilBERT}_4$ & 81.2 & 64.1 & 72.7 \\
$\mbox{TinyBERT}_4$ & 81.0 & 68.2 & 74.6\\
$\bm{\mbox{MLKD-BERT}_4}$ & \textbf{82.0} & \textbf{68.9} & \textbf{75.5}\\
\hline
$\mbox{BERT}_6\mbox{-PKD}$ & 85.3 & 69.8 & 77.6 \\
$\mbox{DistilBERT}_6$ & 86.2 & 69.5 & 77.9 \\
$\mbox{TinyBERT}_6$ & 88.0 & 76.1 & 82.1\\
M\scriptsize{INI}\small LMv2 & - & 76.3 & - \\
$\bm{\mbox{MLKD-BERT}_6}$ & \textbf{88.3} & \textbf{76.5} & \textbf{82.4}\\
\hline
\end{tabular}
}
\end{small}
\caption{ Comparative Studies on SQuAD 1.1 and SQuAD 2.0.}
\label{tab:qa}
\end{table}

\begin{table*}[t]
\centering
\begin{small}
\resizebox{\textwidth}{!}{
\begin{tabular}{l|c|ccc|ccc}
\hline
\multirow{2}*{$\bm{\mbox{MLKD-BERT}_4}$}&\multirow{2}*{$\bm{B_{\mbox{\textbf{\scriptsize{size}}}}}$}&\multicolumn{3}{c|}{\textbf{MNLI-m}}& \multicolumn{3}{c}{\textbf{MNLI-mm}}\\
&&$\bm{A_h^S=12}$&$\bm{A_h^S=6}$&$\bm{A_h^S=3}$&$\bm{A_h^S=12}$&$\bm{A_h^S=6}$&$\bm{A_h^S=3}$\\
\hline
&\multirow{2}*{1}&\multirow{2}*{4.23 $\pm$ 0.19}&4.12 $\pm$ 0.24&4.07 $\pm$ 0.25&\multirow{2}*{4.24 $\pm$ 0.28}&4.14 $\pm$ 0.40&4.03 $\pm$ 0.27\\
&&&(-2.60\%)&(-3.78\%)&&(-2.36\%)&(-4.95\%)\\
&\multirow{2}*{16}&\multirow{2}*{5.46 $\pm$ 0.19}&5.09 $\pm$ 0.22&4.90 $\pm$ 0.32&\multirow{2}*{5.47 $\pm$ 0.17}&5.10 $\pm$ 0.29&4.84 $\pm$ 0.20\\
\textbf{Inference}&&&(-6.78\%)&(-10.26\%)&&(-6.76\%)&(-11.52\%)\\
\textbf{time(ms)}&\multirow{2}*{32}&\multirow{2}*{10.34 $\pm$ 0.18}&9.46 $\pm$ 0.17 &9.00 $\pm$ 0.30&\multirow{2}*{10.35 $\pm$ 0.20}&9.46 $\pm$ 0.13&8.99 $\pm$ 0.24\\
&&&(-8.51\%)&(-12.96\%)&&(-8.60\%)&(-13.14\%)\\
&\multirow{2}*{64}&\multirow{2}*{19.28 $\pm$ 0.20}&17.52 $\pm$ 0.22&16.56 $\pm$ 0.18&\multirow{2}*{19.28 $\pm$ 0.21}&17.49 $\pm$ 0.17&16.59 $\pm$ 0.17\\
&&&(-9.13\%)&(-14.11\%)&&(-9.28\%)&(-13.95\%)\\
\hline
\multirow{2}*{\textbf{Performance}}&&\multirow{2}*{82.0}&81.4&80.6&\multirow{2}*{80.7}&80.6&79.5\\
&&&(-0.73\%)&(-1.70\%)&&(-0.12\%)&(-1.49\%)\\
\hline
\end{tabular}
}
\end{small}
\caption{Effect of Attention Head Number $A_h^S$ on Model Performance and Inference Time.}
\label{tab:tab2}
\end{table*}
In summary, the sample-similarity, sample-contrastive, and soft-label distillation loss functions are summed up to the second stage distillation loss function in Eqn.(\ref{eqa:eqa21}):
\begin{equation}
    \mathcal{L}_{\mbox{\scriptsize{Stage\mbox{\,}2}}}=\mathcal{L}_{\mbox{\scriptsize{SS}}}+\mathcal{L}_{\mbox{\scriptsize{SC}}}+\mathcal{L}_{\mbox{\scriptsize{KD}}}\label{eqa:eqa21}
\end{equation}

\section{Experiments}\label{section:sec5}
\subsection{Experimental Datasets}
Our experiments are conducted on General Language Understanding Evaluation (GLUE) \citep{DBLP:conf/iclr/WangSMHLB19} benchmark and extractive question answering tasks. The former is sentence-level task while the latter is token-level task.
GLUE includes 8 tasks: 
1) Corpus of Linguistic Acceptability (CoLA) \citep{DBLP:journals/tacl/WarstadtSB19}; 
2) Stanford Sentiment Treebank (SST-2) \citep{DBLP:conf/emnlp/SocherPWCMNP13}; 
3) Microsoft Research Paraphrase Corpus (MRPC) \citep{DBLP:conf/acl-iwp/DolanB05}; 
4) Semantic Textual Similarity Benchmark (STS-B) \citep{DBLP:journals/corr/abs-1708-00055};
5) Quora Question Pairs (QQP) \citep{chen2018quora};
6) Question Natural Language Inference (QNLI)
\citep{DBLP:conf/emnlp/RajpurkarZLL16};
7) Recognizing Textual Entailment (RTE) \citep{DBLP:conf/tac/BentivogliMDDG09};
8) Multi-Genre Natural Language Inference (MNLI) \citep{DBLP:conf/naacl/WilliamsNB18}, which is further divided into 
in-domain (MNLI-m) and cross-domain (MNLI-mm) tasks.
The extractive question answering tasks include SQuAD 1.1 \citep{DBLP:conf/emnlp/RajpurkarZLL16} and SQuAD 2.0 \citep{DBLP:conf/acl/RajpurkarJL18}.
The GLUE tasks are evaluated on GLUE test sets and the extractive question answering tasks are evaluated on dev sets.

\subsection{Experimental Setup}
We take a pre-trained language model BERT-base \citep{DBLP:conf/naacl/DevlinCLT19} with 109M parameters as the teacher model (number of layers $M=12$, hidden size $d_h^T=768$, intermediate size $d_h^{T'}=3072$, and attention head number $A_h^T=12$), which is fine-tuned for each specific task. 
Two student models are instantiated for comparative studies: 1) $\mbox{MLKD-BERT}_4$ ($N=4, d_h^S=312, d_h^{S'}=1200, A_h^S=12$) with 14.5M parameters; and 2) $\mbox{MLKD-BERT}_6$ ($N=6, d_h^S=768, d_h^{S'}=3072, A_h^S=12$) with 67.0M parameters.
The student models are initialized with the general distillation model delivered by TinyBERT\footnote{\url{https://github.com/huawei-noah/Pretrained-Language-Model/tree/master/TinyBERT}}. The detailed hyper-parameters are presented in Appendix \ref{appen:appenA}.

\subsection{Comparative Studies}
Comparative studies are conducted to evaluate performance of MLKD-BERT against state-of-the-art BERT distillation baselines, including BERT-PKD \citep{DBLP:conf/emnlp/SunCGL19}, DistilBERT \citep{DBLP:journals/corr/abs-1910-01108}, BERT-EMD \citep{DBLP:conf/emnlp/LiLZXYJ20}, TinyBERT \citep{DBLP:conf/emnlp/JiaoYSJCL0L20}, and  M\small{INI}\normalsize LMv2 \citep{DBLP:conf/acl/WangBHDW21}. 
In addition, to measure the performance improvement delivered by distillation, MLKD-BERT is compared with a same structured pre-trained BERT model $\mbox{BERT}_{\mbox{\scriptsize{TINY}}}$ \citep{DBLP:journals/corr/abs-1908-08962}, which is fine-tuned for each specific task without knowledge distillation.
Note that all experiments are done without data augmentation.

The comparative results evaluated on the test sets of GLUE official benchmark \footnote{\url{https://gluebenchmark.com}} are presented in Table \ref{tab:tab1}. 
Here, different evaluation indices are adopted with regarding to the tasks: F1 metric for MRPC and QQP, Spearman correlation for STS-B, Matthew’s correlation for CoLA, and Accuracy for the other tasks.

Experimental results in Table \ref{tab:tab1} show that, among 4-layer models, our $\mbox{MLKD-BERT}_4$ ranks first on average performance and 5 specific tasks, while ranks second on the rest tasks.
Among 6-layer models, our $\mbox{MLKD-BERT}_6$ delivers similar results.
Moreover, $\mbox{MLKD-BERT}_4$ performs better than $\mbox{BERT}_4\mbox{-PKD}$ and $\mbox{DistilBERT}_4$, with only $30\%$ parameters and inference time. 
Therefore, our MLKD-BERT has an average improved performance, compared to state-of-the-art knowledge distillation methods on BERT.

To evaluate the effectiveness of MLKD-BERT distillation, MLKD-BERT is compared with $\mbox{BERT}_{\mbox{\scriptsize{TINY}}}$ and its teacher BERT-base.
Results show that our $\mbox{MLKD-BERT}_4$ is consistently better than $\mbox{BERT}_{\mbox{\scriptsize{TINY}}}$ on all tasks with $5.3$ improvement on average.
Comparing with teacher model BERT-base, $\mbox{MLKD-BERT}_4$ is 7.5x smaller and 9.4x faster, while keeping average $95.1\%$ performance of its teacher; $\mbox{MLKD-BERT}_6$ keeps average $99.5\%$ performance of its teacher, with only $50\%$ parameters and inference time.
As such, the distillation strategies designed for MLKD-BERT are effective to enhance model performance. 

The comparative results evaluated on the dev sets of SQuAD 1.1 and SQuAD 2.0 are presented in Table \ref{tab:qa}, with F1 metric for evaluation.
As those two tasks are token-level tasks, we remove $\mathcal{L}_{\mbox{\scriptsize{SS}}}$ and $\mathcal{L}_{\mbox{\scriptsize{SC}}}$ in Stage 2. 
The results in Table \ref{tab:qa} show that our $\mbox{MLKD-BERT}_4$ and $\mbox{MLKD-BERT}_6$ outperform all other methods on SQuAD 1.1 and SQuAD 2.0, which further demonstrates the effectiveness of our method.

\subsection{Inference Time vs. Performance} 
As MLKD-BERT can flexibly set attention head number for student model, this group of experiments are to evaluate its effect on model performance and inference time. 
4-layer student model $\mbox{MLKD-BERT}_4$ instantiated with varied numbers of attention heads are experimented on GLUE tasks. $\mbox{MLKD-BERT}_4$ with 12 attention heads is taken as baseline as it has the same number of attention heads as its teacher.
The inference time is measured by the time ($\pm$ standard deviation) required for each batch data on a single GeForce RTX 2080Ti GPU. The batch size $B_{\mbox{\scriptsize{size}}}$ is varied to provide more insights. 

As shown in Table \ref{tab:tab2}, with decrease of attention head number, the inference time decreases very fast with relatively little performance drop. 
And such effect is emphasized with the increase of batch size. As for batch size of 64 on MNLI-m task, when attention head number drops from 12 to 3, the inference time of $\mbox{MLKD-BERT}_4$ decreases 14.11\% while keeping over 98\% prediction performance.
More experimental results presented in Appendix \ref{appen:appenC}, deliver similar effect. 
Therefore, the flexible setting of student attention head number would allow  substantial inference time decrease at little expense of performance drop.

\begin{table}[t]
\centering
\begin{small}
\resizebox{\hsize}{!}{
\begin{tabular}{l|c|ccc|c}
\hline
\multirow{2}*{\textbf{Model}}&\multirow{2}*{$\bm{A_s}$}&\multirow{2}*{\textbf{MNLI-m/-mm}}&\multirow{2}*{\textbf{SST-2}}&\multirow{2}*{\textbf{QQP}}&\multirow{2}*{\textbf{Avg}}\\
&&&&\\
\hline
&3&\textbf{80.6}/\textbf{79.5}&\textbf{91.9}&\textbf{70.0}&\textbf{80.5}\\
$\bm{A_h^S=3}$&6&\textbf{80.6}/79.0&90.4&69.8&79.9\\
&12&80.4/79.3&91.0&69.9&80.1\\
\hline
&6&\textbf{81.4}/\textbf{80.6}&\textbf{92.0}&\textbf{70.4}&\textbf{81.1}\\
$\bm{A_h^S=6}$&8&80.7/80.0&91.3&\textbf{70.4}&80.6\\
&12&81.0/80.2&91.9&\textbf{70.4}&80.9\\
\hline
\end{tabular}
}
\end{small}
\caption{Effect of MHA-split Number on Model Performance.}
\label{tab:tab4}
\end{table}

\begin{table}[t]
\centering
\begin{small}
\resizebox{\hsize}{!}{
\begin{tabular}{l|c|ccc|c}
\hline
\multirow{2}*{\textbf{Model}}&\multirow{2}*{\textbf{Method}}&\multirow{2}*{\textbf{MNLI-m/-mm}}&\multirow{2}*{\textbf{SST-2}}&\multirow{2}*{\textbf{QQP}}&\multirow{2}*{\textbf{Avg}}\\
&&&&\\
\hline
&Concat-split&\textbf{80.6}/\textbf{79.5}&\textbf{91.9}&\textbf{70.0}&\textbf{80.5}\\
$\bm{A_h^S=3}$&Average&80.3/79.2&90.8&\textbf{70.0}&80.0\\
&Random&80.4/79.2&90.8&69.9&80.0\\
\hline
&Concat-split&\textbf{81.4}/\textbf{80.6}&\textbf{92.0}&70.4&\textbf{81.1}\\
$\bm{A_h^S=6}$&Average&81.2/80.4&91.9&\textbf{70.5}&81.0\\
&Random&\textbf{81.4}/80.0&\textbf{92.0}&70.3&80.9\\
\hline
\end{tabular}
}
\end{small}
\caption{Concat-split vs. Average and Random mapping}
\label{tab:tab3}
\end{table}

\subsection{MHA-split Studies}
First we study the effect of MHA-split number $A_s$ on model performance in MHA distillation. 
We instantiate $\mbox{MLKD-BERT}_4$  with 3 attention heads ($A_h^S=3$) and 6 attention heads ($A_h^S=6$), respectively. 
For the student model with $A_h^S=3$, the number of MHA-splits varies with 3, 6 and 12; and for the model with $A_h^S=6$, the number of MHA-splits varies with 6, 8 and 12. 
The training is conducted under the supervision of teacher model BERT-base, which has 12 attention heads.

Table \ref{tab:tab4} shows that, the student model performs best, when the MHA-split number $A_s$ equals to the number of student attention heads $A_h^S$. 
This might because it could keep the integrity of student model's subspaces.
That explains why we suggest setting the number of MHA-splits as the number of student attention heads.

As the number of attention heads in student model could be smaller than that in teacher model, the teacher attention heads are divided into several MHA-splits by the number of student attention heads, so that each student attention head is mapped to several teacher attention heads in a MHA-split.
Our method (named Concat-split) concatenates all teacher attention heads in a MHA-split together. 
We could have two other methods to map student attention head to its teacher in a MHA-split: 1) Average mapping averages the teacher attention heads in a MHA-split; 2) Random mapping randomly selects one teacher attention head in a MHA-split.

The second group of experiments are conducted to compare the performance of Concat-split against Average and Random mapping.
$\mbox{MLKD-BERT}_4$ with 3 attention heads ($A_h^S=3$) and 6 attention heads ($A_h^S=6$) are instantiated, respectively.
As shown in Table \ref{tab:tab3}, our Concat-split outperforms the two other methods on almost all tasks. 
We think it is because our method has less information loss.

\begin{table}[t]
\centering
\begin{small}
\resizebox{\hsize}{!}{
\begin{tabular}{l|ccc|c}
\hline
\multirow{2}*{\textbf{Model}}&\multirow{2}*{\textbf{MNLI-m/-mm}}&\multirow{2}*{\textbf{CoLA}}&\multirow{2}*{\textbf{MRPC}}&\multirow{2}*{\textbf{Avg}}\\
&&&&\\
\hline
\multirow{2}*{$\mbox{MLKD-BERT}_4$}&\multirow{2}*{\textbf{82.0}/80.7}&\multirow{2}*{\textbf{35.5}}&\multirow{2}*{\textbf{86.3}}&\multirow{2}*{\textbf{71.2}}\\
&&&&\\
\hline
$\mathcal{L}_{\mbox{\scriptsize{FFN}}}$ for both FFN &\multirow{2}*{81.7/\textbf{80.8}}&\multirow{2}*{25.7}&\multirow{2}*{85.7}&\multirow{2}*{68.5}\\
and MHA sub-layers&&&&\\
$\mathcal{L}_{\mbox{\scriptsize{MHA}}}$ for both FFN&\multirow{2}*{80.6/79.7}&\multirow{2}*{26.6}&\multirow{2}*{85.6}&\multirow{2}*{68.1}\\
and MHA sub-layers&&&&\\
\hline
\end{tabular}
}
\end{small}
\caption{FFN vs. MHA Distillation.}
\label{tab:tab6}
\end{table}

\subsection{FFN vs. MHA Distillation}
In Transformer-layer distillation, as FFN distillation and MHA distillation can be applied on both FFN sub-layer and MHA sub-layer, we are to study whether those two distillations could replace each other.
As shown in Table \ref{tab:tab6}, neither MHA distillation $\mathcal{L}_{\mbox{\scriptsize{MHA}}}$ nor 
FFN distillation $\mathcal{L}_{\mbox{\scriptsize{FFN}}}$ could perform well on both FFN and MHA sub-layers. 
Instead, $\mathcal{L}_{\mbox{\scriptsize{MHA}}}$ on MHA sub-layer jointing $\mathcal{L}_{\mbox{\scriptsize{FFN}}}$ on FFN sub-layer delivers best performance, which is conducted by our method $\mbox{MLKD-BERT}_4$.

Since the outputs of FFN sub-layer contain feature representations of tokens while the outputs of MHA sub-layer encode the relations of different tokens, feature-level distillation (FFN distillation) fits better for FFN sub-layer, and relation-level distillation (MHA distillation) performs better for MHA. That might explain why our method performs best.

\subsection{Ablation Studies}
The effect of different distillation loss functions (including $\mathcal{L}_{\mbox{\scriptsize{EMB}}}$, $\mathcal{L}_{\mbox{\scriptsize{MHA}}}$, $\mathcal{L}_{\mbox{\scriptsize{FFN}}}$, $\mathcal{L}_{\mbox{\scriptsize{SS}}}$, and $\mathcal{L}_{\mbox{\scriptsize{SC}}}$)
on model performance are evaluated by ablation studies on MNLI-m/-mm, CoLA and MRPC tasks.
$\mbox{MLKD-BERT}_4$ keeping all distillation loss functions is taken as the baseline.

As shown in Table \ref{tab:tab5}, greater drop in performance indicates more importance of corresponding distillation loss function. 
As such, according to average performance, the distillation loss functions $\mathcal{L}_{\mbox{\scriptsize{FFN}}}$, $\mathcal{L}_{\mbox{\scriptsize{MHA}}}$, $\mathcal{L}_{\mbox{\scriptsize{SC}}}$, $\mathcal{L}_{\mbox{\scriptsize{SS}}}$, and $\mathcal{L}_{\mbox{\scriptsize{EMB}}}$ are listed in importance descending order.  
As the removal of any distillation loss function leads to average performance drop, we may conclude that the distillation loss functions proposed by MLKD-BERT are all effective to enhance performance. 
Therefore, the relation-level knowledge ($\mathcal{L}_{\mbox{\scriptsize{EMB}}}$, $\mathcal{L}_{\mbox{\scriptsize{MHA}}}$, $\mathcal{L}_{\mbox{\scriptsize{SS}}}$, $\mathcal{L}_{\mbox{\scriptsize{SC}}}$) can be complementary to feature-level knowledge ($\mathcal{L}_{\mbox{\scriptsize{FFN}}}$) for performance enhancement.
That could explain why our MLKD-BERT outperforms state-of-the-art knowledge distillation methods on BERT. 

\begin{table}[t]
\centering
\begin{small}
\resizebox{\hsize}{!}{
\begin{tabular}{l|ccc|c}
\hline
\multirow{2}*{\textbf{Model}}&\multirow{2}*{\textbf{MNLI-m/-mm}}&\multirow{2}*{\textbf{CoLA}}&\multirow{2}*{\textbf{MRPC}}&\multirow{2}*{\textbf{Avg}}\\
&&&&\\
\hline
$\mbox{MLKD-BERT}_4$&82.0/80.7&35.5&86.3&71.2\\
\hline
w/o $\mathcal{L}_{\mbox{\scriptsize{EMB}}}$&81.9/80.8&33.3&86.9&70.7\\
w/o $\mathcal{L}_{\mbox{\scriptsize{MHA}}}$&81.8/80.3&\textbf{31.3}&86.1&\textbf{69.9}\\
w/o $\mathcal{L}_{\mbox{\scriptsize{FFN}}}$&81.7/\textbf{80.1}&32.7&\textbf{84.9}&\textbf{69.9}\\
w/o $\mathcal{L}_{\mbox{\scriptsize{SS}}}$&81.8/80.4&33.7&85.2&70.3\\
w/o $\mathcal{L}_{\mbox{\scriptsize{SC}}}$&\textbf{81.6}/80.4&32.2&86.2&70.1\\
\hline
\end{tabular}
}
\end{small}
\caption{Effect of Distillation Loss Functions on Performance.
The worst performance on each task is in-bold.}
\label{tab:tab5}
\end{table}

\begin{table}[t]
\centering
\begin{small}
\resizebox{\hsize}{!}{
\begin{tabular}{l|ccc|c}
\hline
\multirow{2}*{\textbf{Model}}&\multirow{2}*{\textbf{MNLI-m/-mm}}&\multirow{2}*{\textbf{CoLA}}&\multirow{2}*{\textbf{MRPC}}&\multirow{2}*{\textbf{Avg}}\\
&&&&\\
\hline
$\mbox{MLKD-BERT}_4$&\multirow{2}*{\textbf{82.0}/\textbf{80.7}}&\multirow{2}*{\textbf{35.5}}&\multirow{2}*{\textbf{86.3}}&\multirow{2}*{\textbf{71.2}}\\
(two-stage)&&&&\\
$\mbox{MLKD-BERT}_4$&\multirow{2}*{80.7/80.2}&\multirow{2}*{28.5}&\multirow{2}*{86.2}&\multirow{2}*{68.9}\\
(one-stage)&&&&\\
\hline
\end{tabular}
}
\end{small}
\caption{One-stage vs. Two-stage Distillation.}
\label{tab:tab7}
\end{table}

\subsection{One-stage vs. Two-stage Distillation}
Here we are to study whether the distillation procedure of MLKD-BERT should be partitioned into two stages.
As we have got 6 distillation loss functions for MLKD-BERT, the one-stage procedure is designed to minimize the sum of the 6 distillation loss functions. 
Experimental results with $\mbox{MLKD-BERT}_4$ on MNLI-m/-mm, CoLA and MRPC tasks are presented in Table \ref{tab:tab7}. We find that $\mbox{MLKD-BERT}_4$ with two-stage distillation outperforms one-stage distillation on all tasks. 
That might because our procedure partition could emphasize different distillation objectives for Stage 1 and Stage 2. Stage 1 emphasizes distilling feature representation and transformation, while Stage 2 emphasizes distilling sample prediction. 

\section{Conclusion}\label{section:sec6}
In this paper, we propose a novel two-stage distillation method MLKD-BERT to distill multi-level knowledge in teacher-student framework. MLKD-BERT enhances existing knowledge distillation methods on BERT in two ways: bringing in valuable relation-level knowledge, and making flexible setting of student attention head number. Experimental results show that MLKD-BERT outperforms state-of-the-art BERT distillation methods on GLUE benchmark and extractive question answering tasks.
We believe that, the easy adaption of our method would be helpful to other Transformer-based PLM compression in teacher-student framework. 

\section*{Limitations}
Our MLKD-BERT has two limitations: 1) The two-stage distillation costs relatively more training time than one-stage methods; 2) MLKD-BERT is limited to handle natural language understanding tasks.

\bibliography{custom}
\bibliographystyle{acl_natbib}

\appendix
\section{Hyper-parameters for Two-stage Distillation}\label{appen:appenA}

\begin{table*}[ht]
\centering
\begin{small}
\resizebox{\textwidth}{!}{
\begin{tabular}{l|cccc|cccccc}
\hline
\multirow{2}*{\textbf{Task}} & \multicolumn{4}{c|}{\textbf{Stage\,1}} & \multicolumn{6}{c}{\textbf{Stage\,2}}\\
\cline{2-11}
&\textbf{Epochs}&\textbf{Batch size}&\textbf{Max seq length}&\textbf{Learning rate}&\textbf{Epochs}&\textbf{Batch size}&\textbf{Max seq length}&\textbf{Learning rate}&$\bm{\rho}$&$\bm{\tau}$\\
\hline
CoLA	&50	&32	&64	&1e-5	&30	&32	&64	&1e-5 &0.07 &1.0\\
MNLI	&6	&32	&128 &3e-5	&6 &32 &128 &3e-5  &0.07 &1.0\\
MRPC	&20	&32	&128 &2e-5	&15	&32	&128 &2e-5 &0.07 &1.0\\
SST-2	&15	&32	&64	&2e-5   &10	&32	&64	&2e-5 &0.07 &1.0\\
STS-B	&20	&32	&128 &3e-5	&15	&32	&128 &3e-5 &0.07 &1.0\\
QQP	    &6	&32	&128 &2e-5	&6	&32	&128 &2e-5 &0.07 &1.0\\
QNLI	&10	&32	&128 &2e-5	&10	&32	&128 &2e-5 &0.07 &1.0\\
RTE	    &20	&32	&128 &2e-5	&15	&32	&128 &2e-5 &0.07 &1.0\\
SQuAD 1.1 &4 &16 &384 &3e-5 &3 &16 &384 &3e-5 &0.07 &1.0\\
SQuAD 2.0 &4 &16 &384 &3e-5 &3 &16 &384 &3e-5 &0.07 &1.0\\
\hline
\end{tabular}
}
\end{small}
\caption{Hyper-parameters for Two-stage Distillation.}
\label{tab:tab8}
\end{table*}

\begin{table*}[t]
\centering
\begin{small}
\resizebox{\textwidth}{!}{
\begin{tabular}{c|c|ccc|ccc}
\hline
\multirow{2}*{$\bm{\mbox{MLKD-BERT}_4}$}&\multirow{2}*{$\bm{B_{\mbox{\textbf{\scriptsize{size}}}}}$} & \multicolumn{3}{c|}{\textbf{QQP}} & \multicolumn{3}{c}{\textbf{SST-2}}\\
&&$\bm{A_h^S=12}$&$\bm{A_h^S=6}$&$\bm{A_h^S=3}$&$\bm{A_h^S=12}$&$\bm{A_h^S=6}$&$\bm{A_h^S=3}$\\
\hline
&\multirow{2}*{1}&\multirow{2}*{4.27 $\pm$ 0.33}&4.16 $\pm$ 0.19&4.09 $\pm$ 0.20&\multirow{2}*{4.24 $\pm$ 0.22}&4.12 $\pm$ 0.22&4.00 $\pm$ 0.19\\
&&&(-2.58\%)&(-4.22\%)&&(-2.83\%)&(-5.66\%)\\
&\multirow{2}*{16}&\multirow{2}*{5.46 $\pm$ 0.17}&5.06 $\pm$ 0.21&4.93 $\pm$ 0.21&\multirow{2}*{5.49 $\pm$ 0.20}&5.07 $\pm$ 0.24&4.91 $\pm$ 0.30\\
\textbf{Inference}&&&(-7.33\%)&(-9.71\%)&&(-7.65\%)&(-10.56\%)\\
\textbf{time(ms)}&\multirow{2}*{32}&\multirow{2}*{10.35 $\pm$ 0.17}&9.48 $\pm$ 0.26&8.99 $\pm$ 0.22&\multirow{2}*{10.39 $\pm$ 0.22}&9.49 $\pm$ 0.18&9.01 $\pm$ 0.24\\
&&&(-8.41\%)&(-13.14\%)&&(-8.66\%)&(-13.28\%)\\
&\multirow{2}*{64}&\multirow{2}*{19.28 $\pm$ 0.21}&17.49 $\pm$ 0.18&16.60 $\pm$ 0.22&\multirow{2}*{19.27 $\pm$ 0.20}&17.49 $\pm$ 0.21&16.59 $\pm$ 0.23\\
&&&(-9.28\%)&(-13.90\%)&&(-9.24\%)&(-13.91\%)\\
\hline
\multirow{2}*{\textbf{Performance}}&&\multirow{2}*{70.6}&70.4&70.0&\multirow{2}*{91.9}&92.0&91.9\\
&&&(-0.28\%)&(-0.85\%)&&(+0.11\%)&(-0.00\%)\\
\hline
\end{tabular}
}
\end{small}
\caption{Effect of Attention Head Number $A_h^S$ on Model Performance and Inference Time on more GLUE tasks.}
\label{tab:tab9}
\end{table*}
The detailed hyper-parameters for two-stage distillation of MLKD-BERT on GLUE tasks and extractive question answering tasks are presented in Table \ref{tab:tab8}.

\section{Effect of Attention Head Number $\bm{A_h^S}$ on Model Performance and Inference Time on more GLUE tasks}\label{appen:appenC}
The effect of attention head number $A_h^S$ on model performance and inference time on more GLUE tasks are summarized in Table \ref{tab:tab9}. We can observe the similar trends as the results on MNLI task in Table \ref{tab:tab2}.

\section{GLUE Dataset}
In this section, we provide a brief description of the tasks in GLUE benchmark \citep{DBLP:conf/iclr/WangSMHLB19}.\\
\textbf{MNLI.} Multi-Genre Natural Language Inference is a large-scale, crowd-sourced entailment classification task \citep{DBLP:conf/naacl/WilliamsNB18}. Given a pair of $\langle{premise,hypothesis}\rangle$, the goal is to predict whether the
$hypothesis$ is an entailment, contradiction, or neutral with respect to the $premise$.\\
\textbf{QQP.} Quora Question Pairs is a collection of question pairs from the website Quora. The task is to determine whether two questions are semantically equivalent \citep{chen2018quora}.\\
\textbf{QNLI.} Question Natural Language Inference is a version of the Stanford Question Answering Dataset which has been converted to a binary sentence pair classification task by \citep{DBLP:conf/iclr/WangSMHLB19}. Given a pair of $\langle{question,context}\rangle$, The task is to determine whether the $context$ contains the $answer$ to the question.\\
\textbf{SST-2.} The Stanford Sentiment Treebank is a binary single-sentence classification task, where the goal is to predict the sentiment of movie reviews \citep{DBLP:conf/emnlp/SocherPWCMNP13}.\\
\textbf{CoLA.} The Corpus of Linguistic Acceptability is a task to predict whether an English sentence is a grammatically correct one \citep{DBLP:journals/tacl/WarstadtSB19}.\\
\textbf{STS-B.} The Semantic Textual Similarity Benchmark is a collection of sentence pairs drawn from news headlines and many other domains \citep{DBLP:journals/corr/abs-1708-00055}. The task aims to evaluate how similar two pieces of texts are by a score from 1 to 5.\\
\textbf{MRPC.} Microsoft Research Paraphrase Corpus is a paraphrase identification dataset where systems aim to identify if two sentences are paraphrases of each other \citep{DBLP:conf/acl-iwp/DolanB05}.\\
\textbf{RTE.} Recognizing Textual Entailment is a binary entailment task with a small training dataset \citep{DBLP:conf/tac/BentivogliMDDG09}.

\section{SQuAD}
Stanford Question Answering Dataset (SQuAD) is a reading comprehension dataset, consisting of questions posed by crowdworkers on a set of Wikipedia articles, where the answer to every question is a segment of text, or span, from the corresponding reading passage, or the question might be unanswerable.\\
\textbf{SQuAD 1.1} \cite{DBLP:conf/emnlp/RajpurkarZLL16} contains 100,000+ question-answer pairs on 500+ articles.\\
\textbf{SQuAD 2.0} \citep{DBLP:conf/acl/RajpurkarJL18} combines the 100,000 questions in SQuAD 1.1 with over 50,000 unanswerable questions written adversarially by crowdworkers to look similar to answerable ones.

\end{document}